# Improved Sparse Ising Optimization

Kenneth M. Zick

Northrop Grumman Systems Corporation, Linthicum, MD 21090 USA

kenneth.zick@ngc.com

*Abstract*—Sparse Ising problems can be found in application areas such as logistics, condensed matter physics and training of deep Boltzmann networks, but can be very difficult to tackle with high efficiency and accuracy. This report presents new data demonstrating significantly higher performance on some longstanding benchmark problems with up to 20,000 variables. The data come from a new heuristic algorithm tested on the large sparse instances from the Gset benchmark suite. Relative to leading reported combinations of speed and accuracy (e.g., from Toshiba's Simulated Bifurcation Machine and Breakout Local Search), a proof-of-concept implementation reached targets 2–4 orders of magnitude faster. For two instances (G72 and G77) the new algorithm discovered a better solution than all previously reported values. Solution bitstrings confirming these two best solutions are provided. The data suggest exciting possibilities for pushing the sparse Ising performance frontier to potentially strengthen algorithm portfolios, AI toolkits and decision-making systems.

## I. Introduction

An important class of optimization problem has binary variables, pairwise interactions, and a sparse network structure. We refer to these broadly as sparse Ising problems, covering the related formulations of sparse MaxCut and sparse quadratic unconstrained binary optimization (QUBO). Sparse Ising problems have long been relevant for studying short-range spin glasses in condensed matter physics [1]. Due to disorder and frustration, spin glass-like problems can be difficult to solve. As noted in Ref. [2], "spin glasses are likely the hardest simple binary optimization problem…they are perfectly suited for benchmarking any new computing paradigm aimed at minimizing a binary quadratic cost function." Beyond lattice structures, sparse Ising problems can arise with random connectivity, small-world connectivity, and other structures. For instance, sparse Ising formulations of independent set problems defined on spatial networks (aka geometric graphs) have recently been proposed for applications such as maximization of airspace, strategic placement of resources, and network design [3]. Such problems are being targeted by classical heuristics and exact solvers [4] as well as analog-mode neutral atom quantum computing [3][5]. In principle, opportunities for sparse Ising formulations abound due to the many sparse networks encountered in the real world, from social networks to the spatial networks found in infrastructure (transportation, power, water, communications), multi-vehicle systems, and biology (protein networks, neural networks, etc.). Sparse Ising problems also have connections to machine learning applications. For instance, a sparse Ising solver was recently applied to training of deep Boltzmann networks [6].

The required level of solution accuracy in binary optimization can vary by mission or even *during* a mission. Thus, solver portfolios may need the ability to trade off other performance dimensions (e.g., speed) in order to hit a range of accuracies. Unsurpassed accuracy for instance may be needed in dynamic competitive scenarios in which the slightest advantage can be extremely valuable or even decisive. Many existing and proposed heuristic solvers, however, appear to have a relatively low ceiling on accuracy for sparse problems. A potentially important advance would be sparse Ising solvers with state-of-the-art speed, power consumption and energy usage for a range of accuracies extending all the way up toward optimal.

After briefly covering some context and background, this report presents new characterization data demonstrating better-than-previously-reported performance on some longstanding sparse Ising benchmark problems. The data come from runs of a proof-of-concept implementation of a novel heuristic algorithm. While the algorithmic innovations and details are to be discussed separately, this report provides initial performance data and comparisons. On the benchmark sparse problems, the previously reported times-to-solution are improved upon by orders of magnitude using the new algorithm paired with just MATLAB and a laptop computer. For two problem instances the new algorithm discovered a better solution than the best previously reported values; solution bitstrings for these new bests are provided.

## II. PRELIMINARIES / BACKGROUND

### A. Benchmark Problems

This report focuses on performance on a set of challenging sparse Ising problem instances that have been used widely for benchmarking new solvers. The set consists of the seven largest instances from Gset MaxCut suite [7], available since 2000 [8]. These seven are selected due to their sparsity (sparse random or square grid structure) and the fact that they have been very difficult to solve to near optimality and can serve as discriminating tests of solver accuracy. Six of the instances have random ±1 weights and a toroidal square grid structure; they are formulated as weighted MaxCut problems and represent 2D Ising (Edwards-Anderson) spin glasses. Note that the toroidal structure makes finding a ground state NP-hard [9]. The remaining instance (G70) represents an unweighted MaxCut problem on a sparse random graph. The instance characteristics are summarized in Table I.

TABLE I
CHARACTERISTICS OF THE LARGEST GSET PROBLEM INSTANCES

| Gset problem instance | Number of variables | Number of edges | Problem Type |
|---|---|---|---|
| G65 | 8000 | 16,000 | Spin glass on toroidal square grid |
| G66 | 9000 | 18,000 | Spin glass on toroidal square grid |
| G67 | 10,000 | 20,000 | Spin glass on toroidal square grid |
| G70 | 10,000 | 9999 | Unweighted MaxCut on sparse random graph |
| G72 | 10,000 | 20,000 | Spin glass on toroidal square grid |
| G77 | 14,000 | 28,000 | Spin glass on toroidal square grid |
| G81 | 20,000 | 40,000 | Spin glass on toroidal square grid |

### B. Metrics

A commonly used measure of heuristic solver speed is time-to-target (TTT) which represents the estimated time to reach a solution equal to or better than a specified target with at least 99% probability. It can be estimated with

$$r = \max\left(1.0, \frac{log(1-.99)}{log(1-P_s)}\right) \quad (1)$$

$$TTT(target) = t_{trial}\, r \quad (2)$$

where $r$ is the number of repetitions (required here to be at least 1.0), $P_s$ is the success probability across multiple trials and $t_{trial}$ represents the execution time per trial. For a given target, the ratio of TTT values associated with two solvers can be used as a measure of relative speed. A further key metric is solution accuracy (aka quality), taken to be the ratio of a solution value vs. the best-known solution value.

### C. Previous Work

A heuristic algorithm named Breakout Local Search (BLS) demonstrated unprecedented solution quality on the seven largest Gset benchmark instances in 2013 [10]. BLS works by alternating between a local search process and a feedback-driven process for jumping toward a new local optimum. A C++ implementation reached its best solution after average times ranging from 1 to 6 hours; TTT values were not provided but estimates are given in Appendix I of this report. A solver approach employing a team of communicating Global Equilibrium Search (GES) processes obtained new best solutions for all seven relevant Gset instances in 2015 [11]. Though TTT was not calculated, two rough estimates are possible using the limited amount of provided data: 82 hours for a G77 solution of 9936 (Team3 $P_s$= 0.05, $t_{trial}$ ~3293.55 s) and 77 hours for a G81 solution of 14046 (Team1 $P_s$= 0.05, $t_{trial}$ ~3082.6 s). A subsequent study employed GES in an experiment involving 20 runs of 20 hours each, during which a new best solution was found for G81 (14056) [12]. Approximation algorithms provide a theoretically guaranteed lower bound on accuracy but have not been competitive with the best heuristics in reaching very high accuracies. A recent approximation algorithm for instance obtained approximation ratios on four of the seven large Gset instances; the ratios were roughly 94% which the authors noted was "a comparably low approximation ratio" relative to other instances [13]. Regarding exact solvers, CPLEX matched the best reported solution on instance G65 with a TTT of 3.4 hours, and also was one of the earliest (if not the first) to reach the latest best-known solution of a smaller instance named G62 [14]. A 2023 pre-print stated that exact solver Gurobi was able to establish optimal solutions for the Gset toroidal square grid problems [15]; solution values were not reported. Toshiba introduced a novel heuristic algorithm called Simulated Bifurcation Machine (SBM) inspired by adiabatic evolution [16]. A recent survey compared the performance of a wide variety of Ising solvers and Ising machine concepts and found that, on fully connected spin glasses with up to 1024 variables, SBM was clearly the fastest [17]. That survey did not identify a state-of-the-art solver for general sparse Ising problems, however. A GPU-based implementation of SBM was tested on the largest Gset instances and was found to reach solution qualities of 99.5 to 99.8% with TTT ranging from 1.5 to 17 hours. In July 2023 the best reported solution to instance G70 was extended from 9591 to 9595 by Ref. [18]; the solution was attained after a long running time of unspecified duration.

Several other heuristic approaches do not yet have extensive Gset results but are contenders for sparse Ising optimization. One is parallel tempering Monte Carlo [19] which simulates replicas of a system at different temperatures and exchanges replicas across temperatures. A version of parallel tempering was accelerated with an application-specific integrated circuit in the 2nd generation Fujitsu Digital Annealer. Due to hardware limitations, it could not solve the six largest Gset problems. On instance G65, it reached a solution quality of 99.89% (5556/5562) with a TTT of 7 hours [14]. A technique that can be combined with parallel tempering is replica cluster moves—two replicas at the same temperature are compared and a cluster of spins is identified, after which the sign of the cluster is changed in each replica. Such moves randomize the configurations and can in some cases greatly accelerate simulations. Examples include Houdayer moves [20] and isoenergetic cluster moves (ICM) [21]. The combined PT+ICM heuristic was found to be faster than simulated annealing and two forms of Digital Annealer on toroidal spin glasses of up to 1024 variables [22]. The ICM technique relies on spin reversal symmetry [21] and presumably is not applicable to problems that have local fields and lack such symmetry. Note that some real-world sparse Ising problems have local fields such as maximum independent set on geometric graphs [4]. Hitachi has put forth a series of hardware-implemented simulated annealing-based solvers called CMOS annealers that use a King's graph lattice [23]. It is not clear how well the solvers can handle toroidal geometries or sparse random graphs, and there has been a paucity of benchmark results. An intriguing new approach is graph neural networks. An initial model was tested on Gset instance G70 [24]; the attained solution quality was quite low in comparison (well below that of BLS) but more advanced models are being developed. A related deep reinforcement learning approach called DIRAC demonstrated capabilities for handling various types of spin glasses of certain sizes [25]. Other recently proposed solvers of note include an opto-electronic solver approach [15] and a sparse Ising solver based on probabilistic bits (pbits) [26].

## III. EXPERIMENTAL RESULTS

*A. Cosm*

A new algorithm named Cosm was recently developed and tested on the largest Gset instances. Cosm is an iterative heuristic algorithm; the specific innovations and details are outside the scope of this initial data report and will be covered elsewhere. A proof-of-concept Cosm solver was implemented with MATLAB (R2020b) and a laptop computer with an Intel Core i7-10850H CPU (6 cores, 2.7 GHz). MATLAB used roughly 5 GB of the available 32 GB of RAM. Each batch of 100 trials was executed by six parallel MATLAB workers. The solver parameter settings were identical for the six spin glass instances and were not tuned for each instance with the exception of one parameter—the number of sweeps per run was tuned for each instance to make use of all six workers and achieve roughly the best TTT. Performance data was collected on the algorithmic metrics of success probability, attained solution accuracy and the number of sweeps (iterations) to various targets. Sweeps-to-target is the number of sweeps per run multiplied by the number of runs from (1). In addition, the speed of the proof-of-concept solver was measured; elapsed times were measured including the outermost loop of trials along with any associated non-trivial pre-processing, enabling calculation of the average $t_{trial}$. TTT was then calculated according to (1) and (2).

*B. Speedups in Reaching 99.5-99.8% Solution Quality*

The speed of the proof-of-concept solver is first compared with the Toshiba SBM GPU implementation which has reported one of the leading combinations of accuracy and speed for the benchmark problems. SBM GPU reached solution qualities of 99.5 to 99.8% on the seven benchmark instances with TTT values of 2–17 hours (Table S4 of Ref. [16]). It turns out that the Cosm solver—intended as a proof-of-concept and not an ideal implementation—is dramatically faster on these instances. It met the SBM targets with a TTT of just 2–12 seconds which is 3–4 orders of magnitude faster than SBM GPU. The results are given in Table I.

In addition to time-to-target, an important metric for heuristic solvers is energy-to-target which combines time and power consumption. Though direct power consumption measurements are not available, the Cosm solver used a laptop CPU with a thermal design power of 45 W [27] while the SBM implementation used an Nvidia GV100 GPU on a Tesla V100 accelerator [Got] with a thermal design power of 300 W [28]. The Toshiba study performed 160 trials in parallel in order to "fully utilize the high computational power of the GPU" [16] and presumably used more than a small fraction of the power budget. Thus, the Cosm solver was not only up to four orders of magnitude faster but likely used less power, implying a significant advantage in energy-to-target.

Note that the $P_s$ values in Table I tended to be slightly above 0.54 since that was a range that made use of all six workers. This is not an indication of the maximum $P_s$. Higher $P_s$ was readily attainable if one were willing to increase the number of sweeps per run. For instance, solving G81 with 6000 sweeps per run instead of 3000 did not improve TTT (~7 s) but did allow $P_s$ to reach 1.0.

TABLE I
SPEEDUP IN REACHING 99.5 TO 99.8% SOLUTION QUALITY

| Gset problem instance | Toshiba SBM GPU results [16] | | | Cosm results | | | | |
|---|---|---|---|---|---|---|---|---|
| | Attained solution | Solution quality vs. new best reported | Time-to-target from Table S4 [16] (s) | Sweeps per run | Success probability $P_s$ (100 runs) | Sweeps-to-target | Time-to-target of proof-of-concept (s) | Speedup of TTT relative to SBM GPU |
| G65 | 5546 | 99.71% | 5651 | 8000 | 0.66 | 34,200 | 2.90 | 1950x |
| G66 | 6342 | 99.65% | 27408 | 8000 | 0.67 | 33,200 | 3.12 | 8780x |
| G67 | 6922 | 99.60% | 6340 | 4000 | 0.57 | 21,800 | 2.27 | 2790x |
| G70 | 9578 | 99.82% | 31599 | 15,000 | 0.65 | 65,800 | 12.2 | 2590x |
| G72 | 6982 | 99.63% | 6142 | 7000 | 0.63 | 32,400 | 3.35 | 1830x |
| G77 | 9904 | 99.64% | 46760 | 7000 | 0.66 | 29,900 | 4.32 | 10,800x |
| G81 | 13992 | 99.54% | 62194 | 3000 | 0.58 | 15,900 | 3.81 | 16,300x |

*C. Other Speed Comparisons*

The 2[nd] generation Fujitsu Digital Annealer reached a solution quality of 99.89% (5556) on instance G65 with TTT = 6.7 hours [14]. The Cosm proof-of-concept reached that target with a TTT of 31.8 s, corresponding to a speedup of 758x. Though just one data point, the fact that a MATLAB and laptop implementation outperformed an ASIC-accelerated parallel tempering solver by such a margin hints at a different class of heuristic performance. Breakout Local Search is less straightforward to compare since TTT was not explicitly reported. Some TTT projections are given in Appendix I, indicating again that the Cosm proof-of-concept reached the targets orders of magnitude faster. Lastly, as mentioned in Secion II.C, Ref. [11] set records corresponding to TTT values of roughly 82 hours for a G77 solution of 9936, and 77 hours for a G81 solution of 14046. The Cosm proof-of-concept reached the same targets in 987 s and 195 s, corresponding to speedups of 300x and 1420x respectively.

*D. Performance at Best Known Solution Quality*

A crucial figure of merit for a solver is attainable solution quality (accuracy), along with the speed and energy efficiency in reaching that solution quality. Cosm was run with an extended number of sweeps in order to determine what solution quality could be attained. The algorithm was run 100 times for each instance using the proof-of-concept solver. The campaigns took just a few hours per instance.

Notably, on two instances Cosm found better solutions than all previously reported solution values. It reached a solution of 7008 on G72 and 9940 on G77. Given that the benchmark problems have been attacked for 23 years, this represents quite a milestone especially for a heuristic solver. Bitstrings for these solutions are provided in Appendices II and III; the solutions are readily verifiable. On two other instances (G70 and G81) Cosm achieved solutions higher than recently believed to be the best known, although at least one other group had found these as well. Cosm reached 14056 on G81, matching a previous result buried in Fig. 3 of Ref. [12]. Cosm reached 9595 on G70 (the sparse random instance), matching a July 2023 finding in Ref. [18]. On the remaining three instances, Cosm matched the best previously reported solution. In all cases Cosm reached the best reported solution multiple times and a TTT estimate could be made, as shown in Table II. For most of these instances this is apparently the first time that TTT is being reported at these solution values.

TABLE II
BEST COSM SOLUTIONS WITH ASSOCIATED PROOF-OF-CONCEPT TIMES-TO-TARGET

| Gset problem instance | Best previous reported solution | Cosm solution/ New best reported solution | Attained solution quality vs. new best | Sweeps per run ($\times 10^6$) | Success probability $P_s$ (100 runs) | Sweeps-to-target ($\times 10^6$) | Time-to-target (s) |
|---|---|---|---|---|---|---|---|
| G65 | 5562 | 5562 | 100% | 0.6 | 0.28 | 8.41 | 744 |
| G66 | 6364 | 6364 | 100% | 1.0 | 0.45 | 7.70 | 755 |
| G67 | 6950 | 6950 | 100% | 0.6 | 0.06 | 44.7 | 4960 |
| G70 | 9595 [1] | 9595 | 100% | 0.4 | 0.02 | 91.2 | 13500 |
| G72 | 7006 [2] | **7008** | 100% | 2.0 | 0.10 | 87.4 | 9690 |
| G77 | 9938 [2] | **9940** | 100% | 2.0 | 0.04 | 226 | 35500 |
| G81 | 14056 [3] | 14056 | 100% | 1.0 | 0.13 | 33.1 | 7750 |

IV. DISCUSSION AND CONCLUSION

The data demonstrate unprecedented heuristic performance on the benchmark problems relative to previously reported results. Note that the proof-of-concept Cosm implementation carried MATLAB overhead and only scratched the surface of parallelization; there are clear paths to more efficient Cosm realizations. Future work should test Cosm on additional sub-classes of sparse Ising problems, assess scalability beyond 20,000 variables, and further characterize its capabilities vs. those of exact solvers. Other recent heuristic solvers should be benchmarked as well, including on the Gset instances. Candidates include PT+ICM [21], DIRAC [25] and pbit-based approaches [26]. Note that in addition to more efficient realizations, there are opportunities for algorithmic enhancements and refinements to Cosm, and for cross-pollination with other approaches.

Given the indications of new performance possibilities, an important future thrust will be identifying, revisiting and developing real-world applications of sparse Ising optimization. Particularly interesting could be connections to sparse neural networks and problems defined on spatial networks [3][4]. Expanding the set of addressable problems through sparsification methods [26] should also be pursued. In any case, the new data reported here suggest exciting possibilities for advancing the frontier of sparse Ising optimization. Heuristic solvers with higher efficiency at the highest levels of accuracy could be useful tools for emerging algorithm portfolios, AI toolkits and decision-making systems.

---

[1] Reported in Ref. [18], July 2023. $P_s$ and execution times were not provided.

[2] Reported in Ref. [11]. Ref. [29] reported a ratio in Table 5.2 implying a better solution was known but specifics were not provided. Ref. [15] reported that a better and optimal solution was established but a solution value was not disclosed.

[3] Reported in Ref. [12], p. 492, Fig. 3 (bottom). $P_s$ and execution times were not provided.

ACKNOWLEDGMENTS

The author would like to thank the following for helpful discussions and collaboration: Northrop Grumman colleagues, Helmut Katzgraber, Nikhil Shukla and Itay Hen.

APPENDIX I: COMPARISON OF COSM PROOF-OF-CONCEPT SOLVER AND BREAKOUT LOCAL SEARCH

Breakout Local Search (BLS) is a heuristic algorithm that set records for accuracy on the Gset benchmark instances in 2013 [10] and is still used today as a comparison point. Time-to-target (TTT) was not explicitly provided in Ref. [10] but is projected from the available data and from equations (1) and (2).[4] Though the amount of BLS data is small, a general picture of the relative performance can be gleaned. The Cosm proof-of-concept solver reached the BLS targets significantly faster, with speedup estimates of at least 2–4 orders of magnitude. Data are given in Table IV. Regarding resources, the proof-of-concept solver used MATLAB 2020b and roughly 5 GB of the available 32 GB of RAM, while the BLS implementation used C++ and apparently had access to 2 GB of RAM [10]. The Cosm proof-of-concept solver used a more modern CPU (Intel Core i7-10850H, 6 cores, 2.7 GHz) with a thermal design power of 45 W [27], while the BLS implementation used a CPU (Intel Xeon E5440 CPU, 4 cores, 2.83 GHz) with a thermal design power of 80 W [30]. Note that in addition to reaching the BLS targets much faster, Cosm attained better solutions in all cases (Section III.D).

TABLE IV
COMPARISON OF BLS AND COSM IMPLEMENTATION SPEED IN REACHING BLS TARGETS

| Gset problem instance | Number of vertices | BLS results from Ref. [10] | | | Projected BLS performance based on Ref. [10] data | | | Cosm proof-of-concept results | |
|---|---|---|---|---|---|---|---|---|---|
| | | Attained solution | Average time to reach attained solution (s) | $P_s$ over 20 runs | Solution quality relative to new best reported solution | Projected average time per run (s) | Projected BLS time-to-target (s) | Time to Ref. [Ben] target (s) | Speedup vs. projected BLS TTT |
| G65 | 8000 | 5558 | 4316 | 0.10 | 99.93% | 431.6 | 18865 | 54.1 | 349x |
| G66 | 9000 | 6360 | 6171 | 0.05 | 99.94% | 308.55 | 27702 | 108 | 257x |
| G67 | 10,000 | 6940 | 3373 | 0.05 | 99.86% | 168.65 | 15142 | 26.4 | 574x |
| G70 | 10,000 | 9541 | 11365 | 0.05 | 99.44% | 568.25 | 51018 | 1.64 | 31,100x |
| G72 | 10,000 | 6998 | 12563 | 0.10 | 99.86% | 1256.3 | 54911 | 51.6 | 1060x |
| G77 | 14,000 | 9926 | 9226 | 0.05 | 99.86% | 461.3 | 41416 | 44.2 | 937x |
| G81 | 20,000 | 14030 | 20422 | 0.05 | 99.82% | 1021.1 | 91676 | 22.9 | 4000x |

---

[4] The BLS success probability $P_s$ is the number of successful runs from Ref. [10] divided by the number of runs (20). The total BLS execution time is assumed here to be the reported average time per success multiplied by the number of successes. The average time per run is then taken to be the total time divided by the number of runs (20).

APPENDIX II: BITSTRING REPRESENTING BEST REPORTED SOLUTION VALUE (7008) FOR GSET INSTANCE G72

Below is a solution bitstring for the Gset G72 problem instance encoding a better solution (7008) than previously reported values in the literature. The solution can be verified by expanding the hexadecimal string to binary values representing variables 1 to 10,000 from left to right, then evaluating the weighted MaxCut objective function using the binary values and the G72 problem graph defined by Ref. [7].

```
AB09C32B897F1290A2B0E080108AAA9980CBD4D26CFB9960B4B3D761A994EAA23298122883DE9C27B5E6160F35D3
55166B94A8B801E4BA42D59005791D977F263CA143895F6A09270BEF3FFF7BB4C116DDDFA2EB1DB38EB5FB03764F8
1463E17829EACE21CC1691087DB098B7163F419A224322673D47638910CDCB7058BEC8E6F92C79EC973EC2ACEFC85
D772F48D3D0BF2D7146B652E3C56E589BE018F696D7FCE01F521B3C7D2590E5DE4D5FAE188D8498A557CA4E6DAE9
537EBA9CA828BF7F6ED87966BF6FD398AE134E479F8FE40CE0215BDECCE9A8FC3A6565FC586BD8F3F910711F2E204
D9B925D26CF94E13D53A387ED8BFA957C6BF50DE81C00E381D3DDAD2C078CAA9247EF3B2BF6041AB9CC6799686F
9A4A30CF0FA04D91699EB25A55A4A05E64B30A067C1F1933E7A5077371020AC85349ECB14447B1BC98D471EDC1379
A1AD56CCCAA5CE9B3FE9E8CC45AE98271B95112A745A5A35351A785E5AB515107D4E85E9E6DF1220C8563A6C6184
55416103BACE415208A6620FFF0E504D0ED87ED3FEEE61E5FE29C436B75D5D0ADE8FD61F7EA5BB7BD91DB742096
D2866363B53A137A5EFD1E7E415CC819C7595192B7A4C133A46E23C60C62E258B0C3AF7463DA1CDEF02943F4126B3
67D9E0D00CA0571E56CD6E24AE3080E649218F4F01C55F64CC24E6C2EE89787A6BD19D2F388D5E79D2628162D3990E
A049C0D85829A0F495CD777E84E589F69BC96F44E7AA1F9F8ED8D7079829111C810CDD439CC47C9F67F2053D3B1C7
1A1EA8D66E5F189462FCA46B5DCF8026EA97D690062483FAD50CD5FEE90A6E50C76FC745B57D28E481C4DAA300A
5E6AA833D6F67C7FC8BC19A90C91334DA0EEC200607B073E1263715D5C7760BFBFB044862E58D14324B81E21C4BAE
DCDE00A2CA35415C2D87BB14218A1670BAF2E55A61D73222E760EC8E9F5991DDFCEAEC9F3AF065387186341D76F6
07794F84CA509C09631DBF8594FA51DEBD7BA68A7E5CCA928CCB9A397E7D78AB96EE808838DFB10FAF50CD96762
C84D472DD5727BC8B7952D23BC5DF853CD43D1FF1FA9FC1478FC67F45A1AC71934121E0BE3AA8001FC1867F48BAF
4FB254797EF9C38AFE9AEEC9F120FC61575692C091A9FB31090A4B91BFED4FB66B2A5574EB0A46F095515BC5722FB
6977D093FCD5483E897B43E55B8702E898163F3E325F89002E3F6184EF770445B3DA7EAB4FC259892BB5497DD1E011D
15015BDFEE60311586E5A4EB98D26E241E2F0C46690D52F10C1C98B3DC9AB09FFE18F9CAC3B70A69725FE176375BC
FF510555F84F44BB54AFC38637F791A0AFD828E233336E3BC1B78F969A888F7FA2963FB025029C1C180DABFCA5A366
C477C907C03C21FB5F80527450A44D84BAE034A1CD6C405452CD303E62E804D99077952844D9264406E63BE1B61CBB
CE4F4B4D825846E559BCCA127B8BA5681D8CAE672C4FED94AE04AECE3B912B901E7C4D48A011CD924489FA119AC
8A05D9605949952542D6171828708333189F0F096AF5535D90075A8406DE0C2B6CE6E1B6C428B344258A63A0240FB0B7
36D9A12152E92A1F37F669447B998CA74322C9FAFA98EA9D7062E6FE5408C69FFF2071AB03C1DCEE4778E9C915FF0
EB8D456CD52DAB467EEF6311A7844CA02B8CA6913366CAC2007577E53F40E096EBEEEF182B19B5E67FF3A5BEA972
87A2047A3212F1EA3F8E07359AB9E6A9AF9D2A663B881E0C9CC21183A7BA003F205095AF95D13C56E2FC46342F
```

APPENDIX III: BITSTRING REPRESENTING BEST REPORTED SOLUTION VALUE (9940) FOR GSET INSTANCE G77

Below is a solution bitstring for the Gset G77 problem instance encoding a better solution (9940) than previously reported values in the literature. The solution can be verified by expanding the hexadecimal string to binary values representing variables 1 to 14,000 from left to right, then evaluating the weighted MaxCut objective function using the binary values and the G77 problem graph defined by Ref. [7].

C4AC172AA225AFFE46F7DC35ADFC8F5DB8D91E006C84C38E866D1E2BB7E91EE8BFE955C956609B12C4094371DC8
451124F96BB456FB8727E327A614F2B539F0627A29BBAE43EFB060EBA3DDA2FDC5085140D6B726F36D61F275B88B8
DCE388C1012660263742C1EF05BA0F8F7610C2D5960B454B3868FDB29A9715082496AEC6AF0CAA679B453480B4F254
CCD18DA206E932CA5FF03A486A05E0971870EFFAC4E101356A6D92FB9539EDEEEFA833692EBDE6B39722DE3BC5F
7D28436EE3A937D7EDE4F26D543BABF97C73BDA5B04F09737865A67A6EE7A90A11BC5BD5EDD4867A5418DB9090B
EBF6710C45DF05B278200DF7CDA8811E274D72B2986438C8AB5767B1875381D06445FF1E7E6C93718C67D2C301A8BE
D7CDECB5AEB154CABF32432CF4DE0DD7FE1601D6BFAB34C94C5444020DE3955D35F6F1301D339825ABB3038921A
186B6244D103606748B9A9714C53EBFFE88D2E608ECC0B118A3288C26CECB390E748F1E327FAE6C891D43FCE5619C
EDE87BCE0D331B840CD22F1D5C77CEC93D71B9F669087176577CE5F11105276B05D5C33AFE25207FA873CB0D07676
6A54FD865048E6CE7A26990F096230CABF506794833F45D054F18BD8DD485800625B8A706C1F54638FFA8FFFEE7D00
AFDE75F19F28D5C09495D040747F09AD826E9C54FAEE0E72D7BB4EB0B290D08FEC748646C8B41CD2C6ECF39BAE1
8ED890B883EB3C7356F08202C31591271BCFCE9BE219AC8A5B3391CBBDD431E6B069B3552AF1D479650724DBE1238
702D132CFA8FB73391013735C19BD44EF7A2BD167B91EAB9A49A6664B3618E899C31BC356517C8B6ACD7FD5FA8FC
09E8C37D6F7F872516867B140AF6431AD402515DD76D4F6B4244CB01A1C0AD93B5EBA642654C400C9A2C180CDC2E
34E7EB8EB6ED7092A9D43BCDAEB6B2327B7176A0987B91AB06A57E67CA15D6DB27367ADA2A1A112B0AC590CBF
F67444DB310E8CA1191093C8283A626E8F2254856BDF1CD3760E330D8D0AB0B734B26449BCDC6EFBE2C3A98FBC45
A4DC65D7C80D193E261F01F895DC168E9657610A14EFDBAF60F5AF4F21FAC39C08A9E10308E20208AC000D8084341
55ECA4C2419F8F2D5A5264EC74FA91978DDB7FDE36963F57A32F2A8DF27EC5916728177E345AE6BCE11BDCECF02A
C874594855790F08278C46D2BDC165A26240512EC205169F04B78C3DE35DD3AD31673D40286283E761BB7FBBFD4DD
7BF481B5A3C0A08CCE8D0152E348B547DC6CF6C928B3E19BE73BA90609F2829261423DA4D09E49073AEBB4AF493E
212B32E89CFBF3628FF33B1FD6EE771EB33C6F917134E3CB88B0E11D9F554E160396863874795875FA3FF5B3F04AE06
D3CE66628E3FB952B04B5647FCE7971350E57E5CE6E75617ED0A741DDB11D019D2558B75F734DAB0A2EDBD88B361
DBCFAAF9591EE93929535E3F905B35CB19F6E46465477C34F886CF1D738E6B87948B5DE629B1B0CA6DD75985BF2B7
6C1330DFB3CCEF79839E568285925DE629FA80C62948B6206304D86E10E5F4622F03446AEA90AFBB22688BB998B9813
06695875093CF7E5FEB128199763FA524D08718A3F9BF5936AD33C56F096DDAB772A8902C2EEDB0A25CF828C85EAF
283531125F50BC082CC2DF6F67EDB0F98B65E7F1321A0EEAAE2C24167409E0BD7F34BF9DED6F5A5A8AC0506EA4D4
F4C21336C85E7D5A604EEA9BA415A605EA00444F78B0E3D50C6A61647AE52B2DE54860A1DC449A1B968096B8D049
1F205DBFFFE1D2A9A162DB4F7AC0BE2217319DB6AFFC7164843D8CA6488925B41ACCE70727D5463BD8B585C406E1
0324AEB2B845CF6AF9DB923A388C7144354C99FD827BA029E8DAB7480D72B24BD01AEC8EC6314ABA6E5B78003FE
A1956F8DA32DA3AB8C7AA331AC33EE483327B7197F9C7637BFAE00683851E8583CDC33CF46695E18C3CD228288E58
78E3B763A73A224D507A4F2C9A88F4C7BFF14B659215763B484BA1535C74D9BBD046F650A93CA43DDDB56751DAD5
2EE42F8EFE463ABC34A1E47B00905F5DE2D77555673269A010CE1677254C94502E95F5ED3315D37749840D9F609075B
639BAA62CC39A67350C5B6A0427789A2EC7D922540EE6A3D9B0B98E496A226F07BBD5D320511782AA65E53BF88450
DAA65C3667BB261BE1EF5E6B88531288C7A791BDAEDC450C9C5788AE9EDC95A96897766068BDADEA47B873898BB
2CB041F419520C9D6D5F65D779567807337690F631E3376D5E98A6AB149D970B0189AB234642DE2D7B5BC2C542D761
EB7420AB3D3C598F572A82A2169575CE74128831E016928BC9326520A7821DA289F1EBEC17BE938C3C66E9C4D1CA4
788EE00EF82AD6A0E374ACB26D7759A41B0CEC6018D6E31E273A1B18EC33BB8FD242A1B857127D955ADAB38BFD9
681B7B7A2F291F579414374D86E93CCBC59E1F67E486A4BC323DFAAEA73AC4F6EC49B2E5B12E289